\documentclass{article}

\usepackage[preprint]{Styles/neurips_2024}

\usepackage[utf8]{inputenc} % allow utf-8 input
\usepackage[T1]{fontenc}    % use 8-bit T1 fonts
\usepackage{hyperref}       % hyperlinks
\usepackage{url}            % simple URL typesetting
\usepackage{booktabs}       % professional-quality tables
\usepackage{amsfonts}       % blackboard math symbols
\usepackage{nicefrac}       % compact symbols for 1/2, etc.
\usepackage{microtype}      % microtypography
\usepackage{xcolor}         % colors
\usepackage{graphicx}
\usepackage{xspace}
\usepackage{enumitem}
\usepackage{mathabx}

\usepackage{wrapfig}

\newcommand{\sysname}{MASAI\xspace}

\newcommand{\datasetsmall}{SWE-bench Lite\xspace}

\newcommand{\sota}{28.33\%\xspace}
\newcommand{\masaicost}{1.96 USD\xspace}

\newcommand{\aider}{Aider\xspace}
\newcommand{\sweagent}{SWE-agent\xspace}
\newcommand{\opendevin}{OpenDevin\xspace}
\newcommand{\acr}{AutoCodeRover\xspace}
\newcommand{\coder}{CodeR\xspace}
\newcommand{\qdev}{Amazon Q-Developer\xspace}
\newcommand{\opencgs}{OpenCGS Starship\xspace}
\newcommand{\mars}{Bytedance MarsCode\xspace}

\newcommand{\moatless}{Moatless\xspace}
\newcommand{\ibm}{IBM Research Agent-101\xspace}

\newcommand{\inp}{\emph{Input}\xspace}

\newcommand{\strategy}{\emph{Strategy}\xspace}

\newcommand{\out}{\emph{Output}\xspace}

\newcommand{\testtemplategen}{Test Template Generator\xspace}
\newcommand{\issuerepro}{Issue Reproducer\xspace}
\newcommand{\localizer}{Edit Localizer\xspace}
\newcommand{\fixer}{Fixer\xspace}
\newcommand{\ranker}{Ranker\xspace}

\newcommand{\react}{ReAct\xspace}

\title{MASAI: Modular Architecture for Software-engineering AI Agents}

\author{%
  Daman Arora$^*$, Atharv Sonwane$^*$, Nalin Wadhwa$^*$ \\
  \textbf{Abhav Mehrotra, Saiteja Utpala, Ramakrishna Bairi} \\
  \textbf{Aditya Kanade, Nagarajan Natarajan} \\
  Microsoft Research India \\ 
  \texttt{\{\href{mailto:daman1209arora@gmail.com}{daman1209arora},
  \href{mailto:atharvs.twm@gmail.com}{atharvs.twm},
  \href{mailto:nalin.wadhwa02@gmail.com}{nalin.wadhwa02}\}@gmail.com}\\
  \texttt{\{\href{mailto:abhavm1@gmail.com}{abhavm1},
  \href{mailto:saitejautpala@gmail.com}{saitejautpala}\}@gmail.com}\\
  \texttt{\{\href{mailto:ram.bairi@microsoft.com}{ram.bairi},
  \href{mailto:kanadeaditya@microsoft.com}{kanadeaditya},
  \href{mailto:nagarajan.natarajan@microsoft.com}{nagarajan.natarajan}
  \}@microsoft.com
  }
}
\begin{document}

\maketitle

\let\thefootnote\relax\footnotetext{$^*$ Equal contribution; names listed in alphabetical order}
\begin{abstract}
A common method to solve complex problems in software engineering, is to divide the problem into multiple sub-problems.
Inspired by this, we propose a Modular Architecture for Software-engineering AI (\sysname) agents, where different LLM-powered sub-agents are instantiated with well-defined objectives and strategies tuned to achieve those objectives. 
Our modular architecture offers several advantages:
(1)~employing and tuning different problem-solving strategies across sub-agents, 
(2)~enabling sub-agents to gather information from different sources scattered throughout a repository, 
and (3)~avoiding unnecessarily long trajectories which inflate costs and add extraneous context.
\sysname enabled us to achieve the highest performance (\sota resolution rate) on the popular and highly challenging \datasetsmall dataset consisting of 300 GitHub issues from 11 Python repositories.
We conduct a comprehensive evaluation of \sysname relative to other agentic methods and analyze the effects of our design decisions and their contribution to the success of \sysname.
\end{abstract}

% \freefootnote{$^*$ Equal contribution; names listed in alphabetical order}

\section{Introduction}
\label{sec:intro}
\begin{wrapfigure}{r}{0.50\textwidth}
  \begin{center}
    \vspace{-20pt}
    \includegraphics[width=0.48\textwidth]{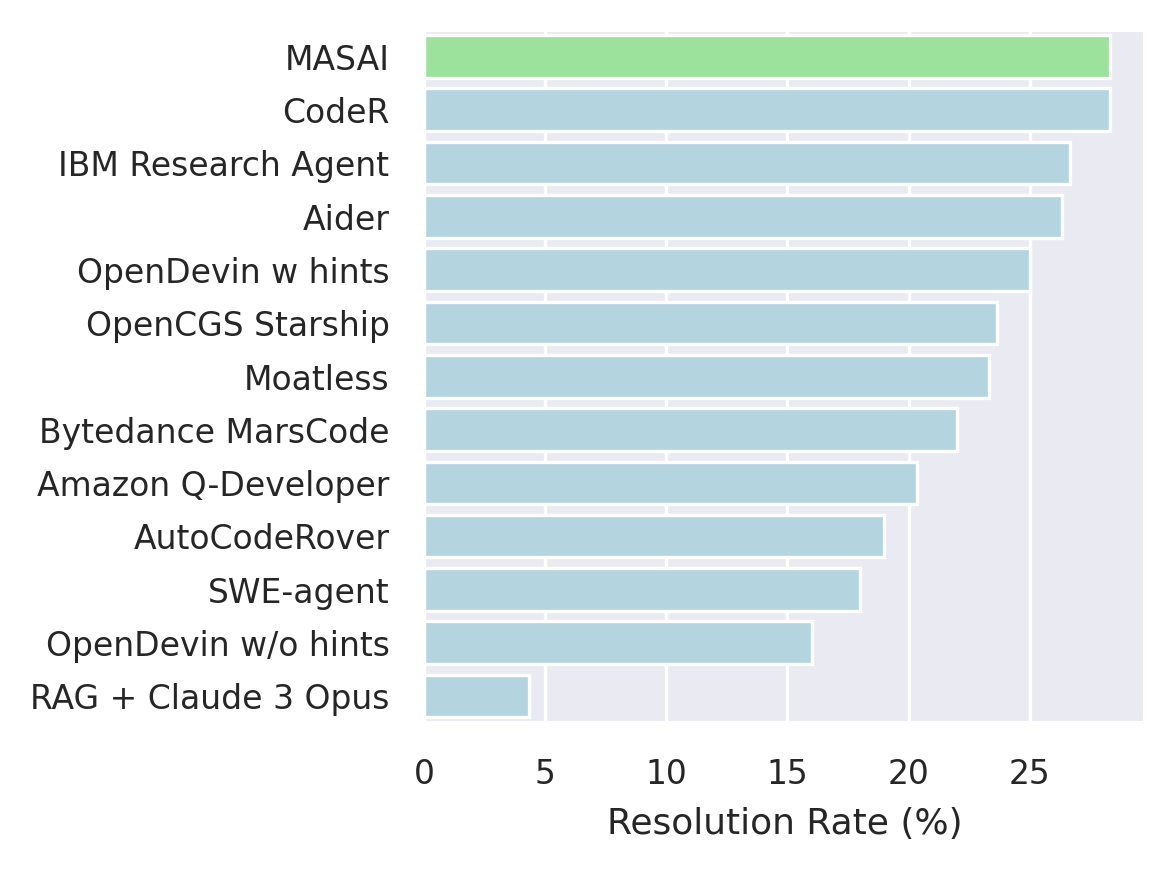}
    \caption{Comparison of \sysname with existing methods.
    \emph{Resolution rate} refers to the percentage of issues in \datasetsmall that are resolved.
    \label{fig:cost_scatter}
    }
    \end{center}
\end{wrapfigure}

Software engineering is a challenging activity which requires exercising various skills such as coding, reasoning, testing, and debugging. The ever growing demand for software calls for better support to software engineers. Recent advances in AI offer much promise in this direction.

Large language models (LLMs) have shown remarkable ability to code~(\citet{chen2021codex,roziere2023code,codegemma}, \emph{inter alia}), reason~\citep{kojima2022large} and plan~\citep{huang2022language}. Iterative reasoning, structured as chains~\citep{wei2022chain} or trees~\citep{yao2024tree} of thought, further enhance their ability to solve complex  problems that require many inter-related steps of reasoning. When combined with tools or environment actions~\citep{yao2023react,patil2023gorilla,schick2024toolformer} and feedback from the environment~\citep{zhou2023language,shinn2024reflexion}, they enable autonomous agents capable of achieving specific goals~\citep{zhang2023igniting}.

As the problem complexity increases, it becomes difficult to devise a single, over-arching strategy that works across the board. Indeed, when faced with a complex coding problem, software engineers break it down into sub-problems and use different strategies to deal with them separately. Inspired by this, we propose a Modular Architecture of Software-engineering AI (\sysname) agents, where different LLM-powered sub-agents are instantiated with well-defined objectives and strategies tuned to achieve those objectives.

Our modular architecture consists of 5 different sub-agents: \textbf{\testtemplategen} which generates a template test case and instructions on how to run it, \textbf{\issuerepro} which writes a test case to reproduce the issue, \textbf{\localizer} which finds files to be edited, \textbf{\fixer} which fixes the issue by generating multiple possible patches, and finally \textbf{\ranker} which ranks the patches based on the generated test. 
When combined, all these individual sub-agents work in tandem to resolve complex real-world software engineering issues.

Our approach offers several advantages: (1)~employing and tuning different problem-solving strategies across sub-agents (e.g., ReAct or CoT), (2)~enabling sub-agents to gather information from different sources scattered throughout a repository (e.g., from a README or a test file), and (3)~avoiding unnecessarily long trajectories which inflate inference costs and pass extraneous context which could degrade performance~\citep{shi2024icml}.

We evaluate \sysname on the popular and highly challenging \datasetsmall dataset~\citep{jimenez2023swe} of 300 GitHub issues from 11 Python repositories. Due to its practical relevance and challenging nature, \datasetsmall has attracted significant efforts from academia, industry and start-ups.
As shown in Figure~\ref{fig:cost_scatter}, with the highest resolution rate of \sota, \sysname achieves state-of-the-art results on \datasetsmall.
The field of AI agents, and specifically software-engineering AI agents, is nascent and rapidly evolving.
 In fact, all the existing methods in Figure~\ref{fig:cost_scatter} have been developed within the past three months. Nevertheless, we do compare against them thoroughly.
 
AI agents for software engineering would encounter many common sub-problems, such as autonomously understanding testing infrastructure and code organization of a repository, writing new tests, localizing bugs, editing large files without introducing syntactic/semantic errors, synthesizing fixes and writing new code.
We believe that it is crucial to understand how different strategies perform on these sub-problems.
Therefore we conduct a thorough investigation into the performance of \sysname and existing methods on \datasetsmall, and present the impact of key design decisions.

In summary, our contributions are:
\begin{enumerate}[leftmargin=20pt,itemsep=2pt]
    \item Propose a modular architecture, \sysname, that allows optimized design of sub-agents separately while combining them to solving larger, end-to-end software engineering tasks.
    \item Show the effectiveness of \sysname by achieving the highest resolution rate on \datasetsmall.
    \item Conduct a thorough investigation into key design decisions of \sysname and the existing methods which can help inform future research and development in this rapidly evolving space.
    \item Contribute our results to the \datasetsmall leaderboard~\citep{masaiPR} for validation.
\end{enumerate}

\section{MASAI Agent Architecture}
\label{sec:design}
Solving a problem in a code repository requires understanding the problem description and the codebase, gathering the necessary information scattered across multiple files, locating the root cause, fixing it and verifying the fix. Instead of treating this as one long chain of reasoning and actions, we propose modularizing the problem into sub-problems and delegating them to different sub-agents.

\begin{figure*}
    \centering
    \includegraphics[width=\linewidth]{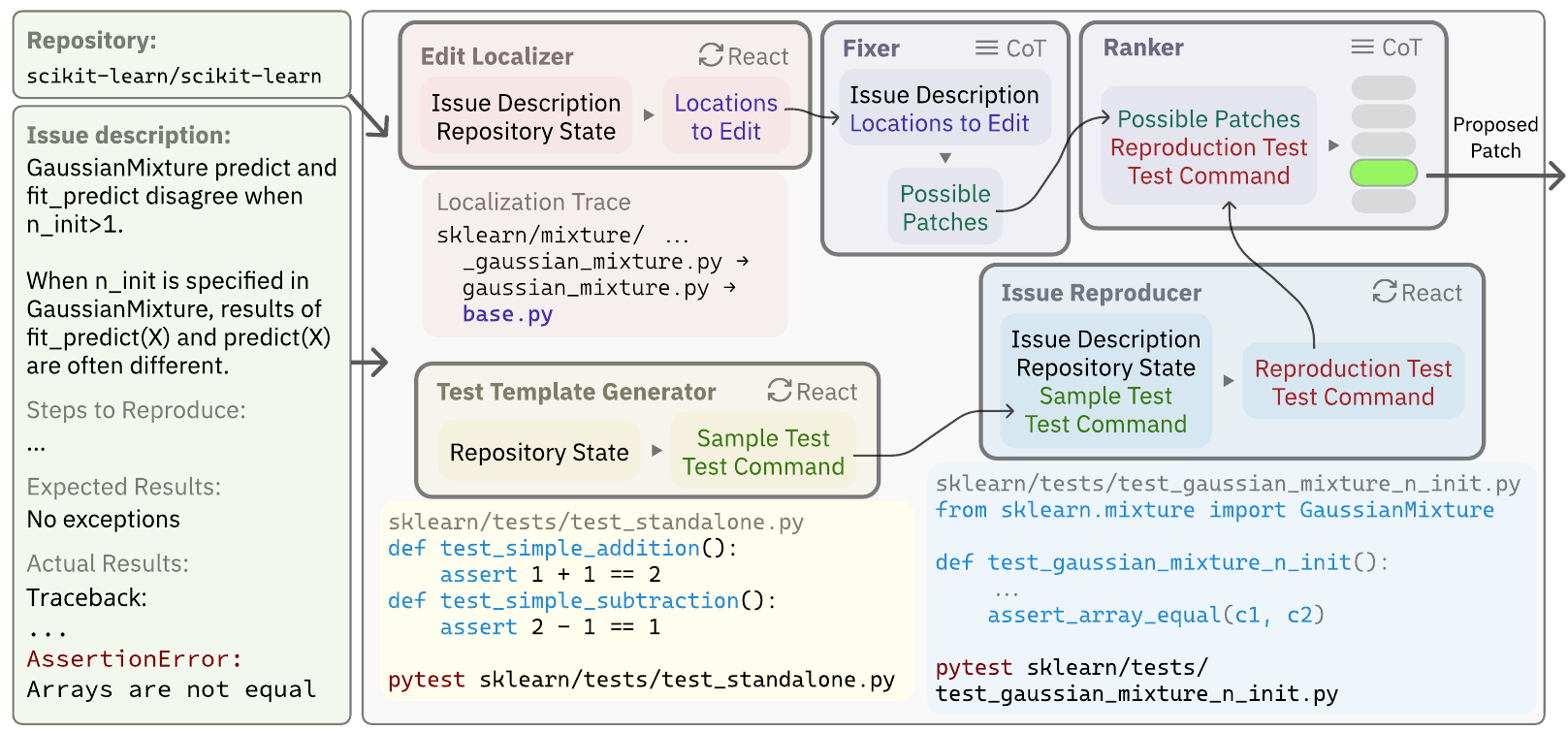}
    \caption{\sysname applied to the task of repository-level issue resolution on an example. \sysname takes a repository and an issue description as input, and produces a single patch. The 5 sub-agents (thick borders) tackle different sub-problems. The information flow between them is shown by directed edges. They marked with the solution strategy and input $\blacktriangleright$ output specification \\ 
    \textit{\textbf{Detail}}: The example issue from the {\tt scikit-learn} repository (id: 13142) describes inconsistent behaviour between two functions relating to {\tt GaussianMixture} and gives an example. The \textbf{\testtemplategen} first generates an example test case along with the command to run the test case. This is given to the \textbf{\issuerepro} which writes a test that reproduces the issue along with the command to run it. The \textbf{\localizer} navigates the repository to find files related to the buggy behaviour of {\tt GaussianMixture} and passes the information to the \textbf{\fixer} which generates a set of possible patches which could fix the issue. Finally, the \textbf{\ranker} takes in the possible patches and the generated reproductions test. Running the tests on the patches, the ranker observes that one of the patches pass while the rest patch fail. It outputs the ranking with the passing patch first which gets selected as the proposed solution.}
    \label{fig:overview}
\end{figure*}

\subsection{Agent Specification and Composition}
A \sysname \emph{agent} is a composition of several \sysname \emph{sub-agents}. A \sysname \emph{sub-agent} is specified by a tuple $\langle \inp, \strategy, \out\rangle$ where

\begin{enumerate}[itemsep=2pt]
    \item \inp to the sub-agent comprises of the code repository, information obtained from other sub-agents as necessary, a set of allowed actions and task instructions.
    \item \strategy is the problem-solving strategy to be followed by the sub-agent in using the LLM to solve its given sub-problem. This could be vanilla completion, CoT~\citep{wei2022chain}, \react~\citep{yao2023react}, RAG~\citep{lewis2020retrieval}, etc.;
    \item \out is the specification of the content that the sub-agent must return upon completion as well the format it must be presented in.
\end{enumerate}

Compared to multi-agent frameworks~\citep{wu2023autogen,qian2023communicative,hong2023metagpt}, the \sysname architecture is simpler, in that, the sub-agents are given modular objectives that do not require explicit one-to-one or group conversations between sub-agents. 
The sub-agents are composed by passing the output from one sub-agent to the input of another sub-agent. 
% This naturally gives rise to a directed graph where if a sub-agent $B$ takes information from a sub-agent $A$ then $A$ must be scheduled before $B$, otherwise, they can run in parallel or sequentially in any order.

\subsection{Action Space}
\label{sec:action-space}

All the sub-agents are presented with a set of actions which allows them to interact with the environment. The actions we use in this work are:

\begin{enumerate}[leftmargin=10pt, itemsep=2pt]

\item {\tt READ(file, class, function)}: Query and read a specific function, class or file. 
All three attributes are not necessary; the agent can specify only a function and a file or even a single file. 
If there exists only one exactly matching code segment with these attributes, then that code is returned.
If there are multiple matches, all their names are returned and the query can be refined if necessary. 
The {\tt READ} action returns a \textbf{lazy representation} that aims to keep the output concise.
When reading a file, only signatures of the top level definitions are presented; when reading a class, the signature of the class (class name and member signatures) are presented and when reading a function, its complete body is presented.

\item {\tt EDIT(file, class, function)}: Marks a code segment for editing. Just like {\tt READ}, this marks a code segment only when a unique match exists. Otherwise, the set of partial matches are returned which may be refined further.

\item {\tt ADD(file)}: Marks a file for code addition. The file must exist for the action to succeed.

\item {\tt WRITE(file, contents)}: Writes the specified content to a file. The specified file can be new or a file that the agent has created earlier.

\item {\tt LIST(folder)}: Lists folder contents if it exists.

\item {\tt COMMAND(command)}: Executes the command in a shell with timeout and truncation of large results.

\item {\tt DONE}: Used by the agent to signal that it has completed its assigned objective.

\end{enumerate}

\subsection{Agent Instantiation}
\label{sec:instantiation}

% \begin{wrapfigure}{r}{0.45\textwidth}
%   \begin{center}
%     \includegraphics[width=0.3\textwidth]{Images/files_touched.png}
%   \end{center}
%   \caption{Repository locations visited by \sysname when solving an issue in {\tt scikit-learn} (ID: 13142). \testtemplategen reads documentation and test files (highlighted in yellow) while \localizer reads files relevant to the issue (highlighted in red). \label{fig:repovisited}}
% \end{wrapfigure}

In this work, we focus on the general task of resolving repository-level issues, as exemplified by the \datasetsmall dataset.
A problem statement consists of an issue description and a repository.
The agent is required to produce a patch so that the issue is resolved.
Issue resolution is checked by ensuring that the relevant, held-out test cases pass.
Below, we refer to \react~\cite{yao2023react} which is a problem-solving strategy that alternates between generating an action to take using an LLM followed by executing the action and using the resulting observations as input for the subsequent action generation.
Chain of Thought (CoT)~\cite{wei2022chain} generates solutions to a problem using an LLM while asking it to generate specific intermediate reasoning steps. 

We instantiate 5 sub-agents to collectively resolve repository-level issues. Figure~\ref{fig:overview} shows the overall architecture of our \sysname agent on a concrete example, along with the information flow between the sub-agents (shown by the solid edges).

\textbf{(1) \testtemplategen}: Discovers how to write and run a new test by analyzing the testing setup specific to the repository.

\begin{itemize}[leftmargin=10pt,itemsep=2pt]
    \item \inp: The repository state (within its execution environment) is provided. {\tt READ}, {\tt LIST}, {\tt COMMAND}, {\tt WRITE} and {\tt DONE} actions are provided.
    \item \strategy: \react.
    \item \out: The code for a template test case (which is issue independent) for the repository along with the command to run it. This is used to aid the \issuerepro sub-agent described next.
    \end{itemize}

\testtemplategen is instructed to explore the documentation and existing tests within the repository to complete its objective and to keep trying until it comes up with a template and a command that passes without exceptions.
\testtemplategen evaluates the output of its \react loop to determine whether the generated test passes without exceptions.
It retries upto a specified limit or until it finds a template that works.

\textbf{(2) \issuerepro}: Writes a test that reproduces the behaviour reported in the given issue.
\begin{itemize}[leftmargin=10pt,itemsep=2pt]
    \item \inp: In addition to the repository state and issue description, the sample test file and the command to run it, generated by the Test Template Generator, are provided. Actions available are {\tt READ}, {\tt LIST}, {\tt COMMAND}, {\tt WRITE} and {\tt DONE}.
    \item \strategy: \react.
    \item \out: The code for a test case which reproduces the issue and would show a change in status (pass vs. fail) when the issue is fixed. It also outputs the shell command to run the test. 
\end{itemize}
\textbf{(3) \localizer}: Navigates the repository and identifies code locations (files, classes, functions) that need to be edited to resolve the issue.
\begin{itemize}[leftmargin=10pt,itemsep=2pt]
    \item \inp: The repository state and the issue description are provided. Available actions are {\tt READ}, {\tt LIST}, {\tt EDIT}, {\tt ADD}, {\tt COMMAND} and {\tt DONE}.
    \item \strategy: \react.
    \item \out: List of code locations (specified through the {\tt EDIT} and {\tt ADD} commands) to edit.
\end{itemize}
If no locations have been marked at the end of the \react loop, then the \localizer selects a set of locations from all of the ones it has read so far.

\textbf{(4) \fixer}: Suggests multiple potential patches to the code locations marked by \localizer that may resolve the issue.
\begin{itemize}[leftmargin=10pt,itemsep=2pt]
    \item \inp: Issue description along with contents of the code locations required to be edited. No actions are given to this sub-agent.
    \item \strategy: CoT.
    \item \out: Multiple possible candidate patches to the provided suspicious code.
\end{itemize}

When prompting the LLM for a possible patch, \fixer asks for the edit in the form of a \textbf{minimal rewrite} instead of rewriting the full sections.
Similar to~\citet{deligiannis2023fixing},
the content of the locations to edit are provided by \fixer with line numbers.
For each edit, the \fixer expects the LLM to output the original version of the code snippet (\textit{pre}) followed by the edited version of this snippet (\textit{post}).
Both these snippets are expected to have a line number for each line. \
\fixer then searches for the \textit{pre} snippet using line numbers in the target file to replace with the \textit{post} version.
If an exact match is not found, it uses \textbf{fuzzy matching} to find the closest matching span for the \textit{pre} snippet.
After replacing with the \textit{post} span, it computes the {\tt diff} of the target file with its contents before the edit. 
Syntactically incorrect edits are rejected and the resultant patches are used downstream.

\textbf{(5) \ranker}: Ranks the candidate patches from the \fixer, using the test generated by  \issuerepro. 
\begin{itemize}[leftmargin=10pt,itemsep=2pt]
    \item \inp: Issue description, candidate patches from \fixer, and the reproduction test (as well as the command to run it) from \issuerepro. No environment actions are allowed.
    \item \strategy: CoT.
    \item \out: Ranking of the candidate patches in the order of likelihood to resolve the issue.
\end{itemize}

For each of the patches,  \ranker first runs the test on each of the patches and  then asks the LLM to determine whether the application of that patch to the repository has caused the provided test to change status (go from failing to passing or vice versa) given the test results.
Based on the output of this, the LLM is then asked to rank the patches.
The top ranked patch is selected as the issue resolution. If the \issuerepro sub-agent could not generate a test, then the Ranker ranks the patches using only the issue description.

\section{Experimental Setup}
\label{sec:setup}
\label{sec:experiments}

\textbf{Dataset}: 
We perform experiments on \datasetsmall~\citep{jimenez2023swe}, a collection of 300 software engineering tasks (predominantly bug fixes) sourced from 11 open-source repositories.
Each task consists of an issue description and the state of the repository on which the issue was raised.
The objective is to produce a patch (that applies to one or more files), which when applied to the repository at the given state, resolves the issue.
The proposed patch for an issue is said to successfully resolve, if the targeted suite of tests, provided as part of the dataset (and revealed only at the time of evaluation), passes on the patched version of the repository
% As stated earlier, we perform experiments on \datasetsmall~\citep{jimenez2023swe} (MIT license).
Each task consists of an issue description and the state of the repository on which the issue was raised.
The objective is to produce a patch given a repository and an issue description, so that the repository after the patch is applied, passes the issue-specific tests (that are never revealed to the agent).

\textbf{Metrics}: We report three metrics: 
(1) \textit{Resolution rate}, the percentage of issues successfully resolved (i.e., pass the issue-specific tests); 
(2) \textit{Localization rate}, the percentage of issues where the patch proposed by a method fully covers the ground-truth patch files, i.e., where recall is 100\% at the file level; and 
(3) \textit{Application rate}, the percentage of issues where the patch proposed by a method successfully applies on the repository (i.e., the Linux command \texttt{patch} does not raise an error).

\textbf{Competing methods}: 
We compare with all the existing methods that are also evaluated on \datasetsmall (with logs \href{https://github.com/swe-bench/experiments/tree/main/evaluation/lite}{here}):

\begin{enumerate}[leftmargin=10pt, itemsep=2pt]

\item \textbf{\sweagent} \citep{yang2024sweagent}: Utilizes a single \react loop along with specialized environment interface with multiple tools. Uses GPT-4 (1106).

\item \textbf{\acr} \citep{zhang2024autocoderover} (ACR): Uses a \react loop for localization and another for generating patches. Uses specialized tools for searching specific code elements (class, method) within other code elements and presenting them as signatures whenever appropriate. Uses GPT-4 (0125).

\item \textbf{\opendevin} \citep{opendevin}: Uses the CodeAct \citep{wang2024executable} framework where the agent (a single \react loop) can execute any bash command along with using various helper commands.
The version of \opendevin with highest reported performance {\tt v1.3\_gpt4o} makes use of {\tt hints\_text} in \datasetsmall, conversation transcript of developers on an issue in GitHub. While we include results from this version, we compare in detail with the highest performing version that does \textit{not} use hints, {\tt v1.5\_gpt4o\_nohints}.

\item \textbf{\aider} \citep{aider}: Uses static analysis to provide a compact view of the repository and, in turn, to determine the file(s) to edit. 
% Uses ReAct loop for editing the identified file(s) until a valid patch that passes \textit{pre-existing} tests is obtained. Uses GPT-4o and Claude 3 Opus on alternate runs.
Limited number of ReAct steps are taken to make an edit to the identified file(s) and iteratively update it until it is syntactically correct and passes existing tests.
After these steps, the final status of linting and pre-existing tests are used to determine whether the tool should be run again from scratch until a plausible solution is found. 
Uses GPT-4o and 365 Claude 3 Opus on alternate runs

\item \textbf{\coder}  \citep{chen2024coder}: 
A multi agent solution with separate agents to reproduce the issue, localize the fault and iteratively edit the code to resolve the issue. 
% A multi-agent solution which reproduces and resolves the issue iteratively.
Uses BM25 along with test coverage statistics for fault localization. Uses GPT-4 (1106).

\item \textbf{\moatless} \citep{moatless}:  Uses a \react loop to localize and another to fix the code. Leverages a semantic search tool that searches with natural language queries for relevant code chunks in the repository.

\item \textbf{RAG}: Uses BM25 to retrieve relevant files which are used to prompt an LLM to generate a patch. We compare with the best-performing RAG model from the \datasetsmall leaderboard~\citep{swebench_leaderboard}: {\tt RAG + Claude 3 Opus}.

\item Along with the above, commercial offerings \textbf{\qdev} \citep{amazonqdev}, \textbf{\mars} \citep{marscode}, \textbf{\opencgs} \citep{opencgs} and \textbf{\ibm} \citep{ibm101} have also reported results on \datasetsmall. 
While we report metrics for these, we are unable to conduct further comparisons with them due to non-availability of detailed logs or any information about their approaches. We do not compare with Devin~\citep{devin} as it reports performance a subset of SWE-bench different from \datasetsmall.
    
\end{enumerate}

\textbf{Implementation}: 
We evaluate \sysname by setting up a fresh development environment with all the requirements and providing the issue description. 
\sysname generates a single patch which is then evaluated using the \datasetsmall testing harness. The {\tt tree-sitter==0.21.1} package is used to implement the lazy representation part of the {\tt READ} function.
We use the GPT-4o model throughout our pipeline.
For \testtemplategen, we start with a temperature of 0 and increase by 0.2 for each attempt.
For \issuerepro, \localizer, and \ranker, we use a temperature of 0; for \fixer, we use 0.5 and sample 5 candidate patches.
We limit the \react loops of the \testtemplategen, \issuerepro, and \localizer to 25 steps and limit \testtemplategen to 3 retries. After the ranker selects the patch, we run an auto-import tool to add missing imports. We discard any edits to pre-existing tests which the agent might have made. The per-issue cost for \sysname is \masaicost on average. We estimate the total cost of our experiments to be $<$10k USD.

\section{Results}
\label{sec:results}
We first present comprehensive results on the \datasetsmall dataset. Then we provide supporting empirical observations and examples that bring out the effectiveness of our design choices.

\subsection{RQ1: Performance on software engineering tasks in \datasetsmall}

We present our main results in Table~\ref{tab:combined_results}. Multiple remarks are in order.

\begin{enumerate}[leftmargin=10pt, itemsep=2pt]
\item Our method, \sysname, achieves the highest resolution rate of 28.33\% on the dataset, thereby establishing a state-of-the-art on the benchmark leaderboard alongside \coder~\citep{masaiPR}.
  
\item Standard RAG baseline (first row) performs substantially poor on the dataset, as has also been established in recent works \citep{jimenez2023swe,chen2024coder}; which is a strong indication of the complexity of the \datasetsmall dataset.

\item \sysname localizes the issue (at a file-level) in 75\% of the cases; the best method in terms of localization rate, \opencgs, at nearly 91\%, however achieves only 23.67\% resolution rate.

\item The (edit) application rate is generally high for all LLM-based agents; \sysname's patches, in particular, successfully apply in over 95\% of the cases.

\end{enumerate}

\subsection{RQ2: Assumptions by different methods} \label{sec:autonomy_rq}

High autonomy and less dependence on external signals (e.g., expert hints) is desirable from software-engineering agents.
In the standard \datasetsmall setup, all agents are provided the issue description along with the repository. 
However, we observe that different methods make different assumptions about available auxiliary information. 

\begin{table}[t]
\centering
\begin{tabular}{l r r r}
\toprule  
% Method&Resolv.&Locl.&Appl.\\
% &rate (\%) & rate (\%)&rate (\%)\\\toprule  
Method&Resolution&Localisation&Application\\
&rate (\%) & rate (\%)&rate (\%)\\\toprule  
RAG&4.33& 48.00& 51.67\\
SWE-agent&18.00 & 61.00& 93.67\\
ACR&19.00 & 62.33& 80.00\\
Q-Dev& 20.33& 71.67& 97.33\\
MarsCode& 22.00& 67.00& 83.67\\
Moatless& 23.33& 73.00& 97.00\\
Starship& 23.67& \textbf{90.67}& \textbf{99.00}\\
OpenDevin&25.00& 77.00& 90.00\\
\ \ \ \ -- hints&16.00& 63.00& 81.33\\
Aider&26.33 & 69.67& 96.67\\
Agent-101& 26.67& 72.67& 97.33\\
CodeR& \textbf{28.33} & 66.67& 74.00\\
\midrule 
\sysname&\textbf{28.33} & 75.00& 95.33\\ \toprule 
\end{tabular}
\caption{Performance of baseline and competing methods on \datasetsmall (best in \textbf{bold}). Our proposed method, \sysname, achieves the best resolution rate (\% issues resolved). Row ``-- hints'' indicates executing \opendevin without using \texttt{hints\_text} in the dataset.}
\label{tab:combined_results}
\end{table}

\begin{itemize}[leftmargin=10pt,itemsep=2pt]

\item All methods apart from RAG and \moatless require that for each task, an environment be set up with the appropriate requirements installed beforehand so that code can be executed.

\item \opendevin avails {\tt hints\_text} provided by \datasetsmall as discussed in Section \ref{sec:experiments}.

\item \aider, when running pre-existing tests, uses pre-determined test commands consist of (1) the testing framework used to run tests in the task repository and (2) specific unit tests that target the code pertaining to the issue at hand. 
The former assumes information about the repository-specific testing framework which is not present in the standard \datasetsmall setup.
In the case of the latter, providing output from only the target test (and not the whole test suite) during \react steps, inadvertently provides additional information about which part of the repository is relevant to the issue. 

\item \coder uses coverage-based code ranking \citep{wong2016survey} for fault localization. As in \aider, this would require repository-specific commands to run pre-existing tests, and instrumentation of the full repository to get coverage information.
However from the available trajectory logs of \coder it does not appear to discover these autonomously.
\end{itemize}

\sysname aims for high autonomy by avoiding dependence on additional inputs, only relying on the original setup proposed by~\citet{jimenez2023swe}.
\sweagent and \acr operate at a similar level of autonomy to \sysname.
Results in Table \ref{tab:combined_results} show that \sysname outperforms all other approaches without making additional assumptions.

\subsection{RQ3: How does \sysname perform effective fault localization from  issue description?}

Localization requires multi-step reasoning to identify the root cause of the error from issue descriptions which are often vague and usually only
describe the problem being observed.
We observe that (1) the choice of \react as the strategy, (2) the specificity of its objective (to only identify files to edit) and (3) the designs of tools available enables the \localizer to perform the required multi-step reasoning in a flexible and robust manner.
Note that (1) and (2) are results of the modularity of \sysname.
\sweagent and \opendevin, methods that do not employ a separate localization sub-agent, achieve 61\% and 63\% localization rates respectively, compared to 75\% achieved by \sysname's \localizer. 

We observe the advantages of using a \react sub-agent, by comparing with \aider which uses a single step CoT approach. 
In the 27 issues solved by \sysname but not by \aider, \aider failed to localize in 10 (37\%) issues whereas among the 21 issues solved by \aider but not by \sysname, \sysname only failed to localize in 3 (14\%) issues.
This shows that better localization plays a role in superior resolution rate.
Comparing the average search steps (as proxy for complexity) required for problems that both \aider and \sysname solved (10.9) and those that only \sysname solved (12.8), we further see that \sysname's \react based \localizer has the flexibility to scale to more complex localization challenges.

[\underline{Example 1}]: \sysname performs \textbf{multi-step reasoning} required for localization in the task \href{https://github.com/scikit-learn/scikit-learn/issues/13070}{\tt scikit-learn\_\_scikit-learn-13142} (described in Fig.~\ref{fig:overview}).
\localizer finds the class mentioned in the issue and then traces symbols and inheritance links to identify the root cause.

[\underline{Example 2}]: The ability of the {\tt READ} action to return \textbf{approximate matches} (Section~\ref{sec:design}) helps in the issue \href{https://github.com/astropy/astropy/issues/14978}{\tt astropy\_\_astropy-14995}.
When the LLM asks for a non-existent {\tt NDDataRef.multiply} method in the {\tt astropy/nddata/nddata.py} file, the action responds with an approximate match {\tt NDArithmeticMixin.multiply} in a different file {\tt astropy/nddata/mixins/ndarithmetic.py}.
Then the sub-agent traces 3 callee links to get to the actual faulty function.

[\underline{Example 3}]: Access to basic \textbf{shell commands} helps the \localizer in the issue \href{https://github.com/matplotlib/matplotlib/issues/25329}{\tt matplotlib\_\_matplotlib-25332}. {\tt grep} is used to look for occurrences of the {\tt FigureBase.\_align\_label\_groups} attribute within the large file {\tt lib/matplotlib/figure.py}.
From the the occurrences (output from {\tt grep}), \sysname finds out that the attribute is set using {\tt cbook.Grouper()} -- the class that needs to be edited to resolve the issue.

Neither \aider nor \coder localized faulty functions correctly in any of the 3 examples.
\opendevin localized Example 2; \sweagent Examples 2 and 3.

\subsection{RQ4: How does \sysname's sampling and ranking compare to iterative repair?}
\label{sec:sampling}

\begin{table}[t]
\centering

\begin{tabular}{l c c}
\toprule 
Selection Strategy &  1 Sample &  5 Samples\\ \toprule  
Oracle&  23.33\%&  35.00\%\\
Random& -& 22.28\%\\
LLM w/o test& -& 23.33\%\\
LLM w/ test (\ranker) & -& 28.33\%\\\toprule 
\end{tabular}
\caption{Resolution rates of \sysname{} on \datasetsmall, with different number of \fixer samples (i.e., candidate patches), using different sample selection strategies (rows, discussed in Section \ref{sec:sampling}).}
\label{tab:pass_at_k}
\end{table}

We observe that sampling multiple repair patches from the \fixer significantly increases the possibility of generating a correct patch, as reported in Table \ref{tab:pass_at_k} (Oracle selection 23.33\% at 1 sample vs 35\% at 5 samples).
However the LLM alone is unable to select amongst theses patches (LLM w/o test). 
This can be overcome by using the output from the generated issue-reproduction test on each patch for ranking the patches (LLM w/ test (\ranker)).

\sysname exploits the above observations through its modularity by (1) leveraging a CoT sampling strategy for \fixer and (2) instantiating independent sub-agents for test generation and repair.
Other methods rely on an iterative approach to extract multiple edits from the LLM asking it to iteratively fix any mistakes it has made.

\begin{table}[t]
\centering
\begin{tabular}{p{2cm} l l l} \toprule
Method&Both&Method&\sysname\\  
% &locl.&resolv.&resolv.\\
&localised&resolved&resolved\\

\toprule 
RAG& 126& 12&\textbf{52} (+ 31.7\%)\\  
ACR&   166&51&\textbf{73} (+ 13.2\%)\\ 
Q-Dev& 191& 55&\textbf{75} (+ 10.5\%)\\ 
SWE-agent&    166&48&\textbf{65} (+ 10.2\%)\\  
Starship& 220& 62&\textbf{81} \phantom{0}(+ 8.6\%)\\ 
OpenDevin&   187&60&\textbf{74} \phantom{0}(+ 7.5\%)\\ 
\ \ -- hints&   164&39&\textbf{67} (+ 17.1\%)\\
Moatless& 193& 62& \textbf{75} \phantom{0}(+ 6.7\%)\\
MarsCode& 182& 59&\textbf{71} \phantom{0}(+ 6.6\%)\\ 
Agent-101& 193& 69& \textbf{72} \phantom{0}(+ 1.6\%)\\
Aider&   189&\textbf{71} &\textbf{71} \phantom{0000\%}(=)\\ 
CodeR&   174&\textbf{77}&72 \phantom{0}\;(- 0.3\%)\\
\toprule
\end{tabular}
\caption{Number of issues resolved by a method (Method resolved) named in the rows and by \sysname (\sysname resolved) among the issues that are successfully localized by both \sysname and the method (``Both localised'' column, out of 300). 
Row-wise max. in bold.}
\label{tab:repair_perf}
\end{table}

We evaluate the benefits of our approach empirically in Table~\ref{tab:repair_perf}.
By controlling for localization, we are comparing the effectiveness of completing the repair. \sysname is substantially more effective at this than most methods, barring \coder and \aider.

As as example, consider the issue \href{https://code.djangoproject.com/ticket/33043}{\tt django\_\_django-14787} where 
\coder, \aider, \opendevin and \sysname all correctly localize the issue, but only \sysname solves it correctly. 
While iterative methods sample one candidate and keep refining it without success, \sysname's \fixer sub-agent generates 5 samples out of which only one is correct -- demonstrating the importance for diverse sampling.
\sysname's \ranker correctly ranks these by utilizing outputs from running the generated reproduction test. \aider submits patch which passes pre-existing tests but is actually incorrect, showing the importance of the generated reproduction test to eliminate false positives.

\subsection{RQ5: How does \sysname perform effective issue reproduction?}

As discussed in the previous RQ, the ability to generate tests that reproduce the stated issue is critical to select \fixer samples. 
Often repositories employ uncommon testing frameworks, that makes this task hard. Consider the issue \href{https://code.djangoproject.com/ticket/32947}{\tt django\_\_django-14672}. This repository proved hard to write tests for since it uses a custom testing framework, which involved having all new test classes derive from a certain base class to run.
\opendevin was unable to reproduce the test; in its attempt to install pytest, it ran out of budget and failed to solve this issue.

To remedy this, we decompose test reproduction into two steps: (1) \testtemplategen reads documentation/existing tests to \textbf{generate a sample test template} and instructions to run; (2) \issuerepro then uses the template as an example to \textbf{create an issue specific test }.
This improves the overall capability of reproducing tests in \sysname, \testtemplategen first goes through the repository, creates a template file that correctly makes use of {\tt django.test.TestCase} to create an example test case as well as the correct command to run it.
The \issuerepro subsequently reproduces the issue correctly, without running into problems that \opendevin faced.

\subsection{RQ6: How does \sysname generate edits that can be applied successfully?}

The representation used to encode edits can have a large impact on the performance.
As discussed in Section~\ref{sec:design}, \sysname prompts the LLM for edits, in the form of a \textbf{minimal rewrite} --- to reproduce the current state of the code snippet it wants to edit, followed by the edited version of this snippet.
Recall that we also employ \textbf{fuzzy matching}  to find the relevant span in the file, by searching for the snippet that best fuzzily matches with the one provided by the model. 
This mitigates copying or line counting mistakes by the LLM, significantly reducing the number of syntax errors introduced when editing.
Our edit representation and fuzzing matching together yield 96.33\% edit application rate  (Table~\ref{tab:combined_results}) which is among the highest.

\section{Related Work}
\label{sec:related}
We have already discussed competing methods evaluated on \datasetsmall, in Sections~\ref{sec:experiments} and \ref{sec:results}.
We now highlight other related work on LLM-powered agents.

\paragraph{Software-engineering agents:} Language Agent Tree Search~\cite{zhou2023language} synergizes reasoning, planning, and acting abilities of LLMs. 
Their strategy relies on determining partial or full termination of the search (e.g., by running provided golden test cases for successful code generation as in HumanEval~\cite{chen2021codex}) and backtracking if necessary; this is often infeasible in complex software engineering tasks we tackle in this paper. 
CodePlan~\citep{bairi2023codeplan} combines LLMs with static analysis-backed planning for repository-level software engineering tasks such as package migration. It relies on compiler feedback and dependency graphs to guide the localization of edits; unlike in our general setting, where the agents are more autonomous, and are equipped to discover localization strategies. 
AlphaCodium~\citep{ridnik2024code} differs from \sysname in that (1) it uses public \textit{and} AI-generated test cases for filtering; (2) is evaluated in the generation (NL2Code) setting.

\paragraph{Conversational and multi-agent frameworks:}
In this line of work~\cite{guo2024large, yang2024if}, (1) the focus is often on the high level aspects of agent design such as conversation protocols. AutoGen~\citep{wu2023autogen} and AgentVerse~\citep{chen2023agentverse} provide abstractions for agent interactions and conversational programming for design of multi-agent systems;  similarly, Dynamic agent networks~\citep{liu2023dynamic} focuses on inference-time agent selection and agent team optimization; and (2) the frameworks are typically instantiated on standard RL or relatively simpler code generation datasets. For instance, AutoDev~\citep{tufano2024autodev} can execute actions like file editing, retrieval, testing, but is evaluated on the HumanEval~\citep{chen2021codex} NL2Code dataset. Similarly, MetaGPT~\citep{hong2023metagpt} and ChatDev~\citep{qian2023communicative}, dialogue-based cooperative agent frameworks, are instantiated on generation tasks involving a few hundred lines of code.

In contrast, we focus on designing a modular agent architecture for solving complex, real-world software engineering tasks, as exemplified by the \datasetsmall dataset.

\paragraph{Divide-and-Conquer approaches:} In this line of work, the given complex task is broken down into multiple sub-goals that are solved individually, and then the solution for the task is synthesized. Multi-level Compositional Reasoning (MCR) Agent~\citep{bhambri2023multi} uses compositional reasoning for instruction following in environments with partial observability and requiring long-horizon planning, such as in robotic navigation. Compositional T2I \citep{wang2024divide} agent uses divide-and-conquer strategy for generating images from complex textual descriptions. SwiftSage~\citep{lin2024swiftsage} agent, inspired by the dual-process theory of human cognition for solving tasks, e.g., closed-world scientific experiments~\citep{wang2022scienceworld}, uses finetuned SLM policy (``Swift'') to decide and execute fast actions, and an LLM (``Sage'') for deliberate planning of sub-goals and for backtracking when necessary.

\section{Conclusions}
\label{sec:conclusions}
As divide-and-conquer helps humans overcome complexity, similar approaches to modularize tasks into sub-tasks can help AI agents as well. In this work, we presented a modular architecture, \sysname, for software-engineering agents. Encouraged by the effectiveness of \sysname on \datasetsmall, we plan to extend it to a larger range of software-engineering tasks, which will also involve building realistic and diverse datasets.

\section{Limitations}
\label{sec:limitations}
Our evaluation is centered on the widely-used \datasetsmall dataset for evaluating software-engineering AI agents. It allowed us to do head-to-head comparison with many agents. However, the breadth of issues covered in \datasetsmall is limited to those that can be validated using tests. In future, we expect us and the community to expand the scope to more diverse issues.

There are a number of LLMs that support code understanding and generation. The modularity of \sysname permits use of different language models in different sub-agents. Due to the time and cost constraints, we have instantiated all sub-agents with GPT-4o. The cost-performance trade-off of using different LLMs and possibly, even small language models (SLMs) is an interesting research problem. The competing methods that we compared against do employ different LLMs, but this still leaves out direct comparison of different LLMs on a fixed solution strategy.

The issue descriptions in \datasetsmall are all in English. This leaves out issues from a large segment of non-English speaking developers. The increasing support for the diverse world languages by LLMs should enable multi-lingual evaluation even in the software engineering domain, which is a problem that we are excited about.

\section{Broader Concerns}
\label{sec:concerns}
Agentic frameworks with the ability to use tools like shell commands can lead to unintended side-effects on the user's system.
Appropriate guardrails and sandboxing can mitigate such problems.

Our approach contributes towards the development of tools to autonomously perform software development tasks.
This raises various security concerns.
The tool may not always follow best practices when writing or editing code, leading to introduction of security vulnerabilities and bugs.
Therefore, it is important for code changes suggested by the tool to be reviewed by expert developers before being deployed to real world systems.

As mentioned in the Section~\ref{sec:limitations}, the dataset we evaluate on (\datasetsmall) as well as the model we use (GPT-4o) are primarily in English.
This limits the usability of our tool to software engineers proficient in English.
Further work is necessary in developing methods for non-English speaking developers in order to prevent this population from being marginalized.

\bibliography{main}
\bibliographystyle{plainnat}
\end{document}